\documentclass[11pt,letterpaper]{article}
\usepackage{emnlp2017}
\usepackage{times}
\usepackage{latexsym}
\usepackage{flushend}

\emnlpfinalcopy

\def\emnlppaperid{11}

\title{Strawman: an Ensemble of Deep Bag-of-Ngrams for Sentiment Analysis}

\author{Kyunghyun Cho\\
  Courant Institute \& Center for Data Science, \\
  New York University \\
  {\tt kyunghyun.cho@nyu.edu}}

\date{}

\begin{document}

\maketitle

\begin{abstract}
This paper describes a builder entry, named ``strawman'', to the sentence-level sentiment analysis task of the ``Build It, Break It'' shared task of the First Workshop on Building Linguistically Generalizable NLP Systems. The goal of a builder is to provide an automated sentiment analyzer that would serve as a target for breakers whose goal is to find pairs of minimally-differing sentences that break the analyzer. 
\end{abstract}

\section{Data and Preprocessing}

\paragraph{Data}

The organizers of the shared task provided two distinct types of training sets. The first set consists of usual sentences paired with their corresponding sentiment labels (+1 for positive and -1 for negative) and confidences (a real value between 0 and 1.) The other set consists of phrases paired similarly with sentiment labels and confidences. In the latter case, the sentiment label may be either -1, 1 or 0 which indicates neutral. There are 6920 sentences and 166,737 phrases.

As the goal of ``strawman'' is to build the most naive and straightforward baseline for the shared task, I have decided to use all the examples from both of the training sets whose sentiment labels were either -1 or 1. In other words, any phrase labelled neutral was discarded. The confidence scores were discarded as well.

The combined data was shuffled first, and then the first 160k examples were used for training and the last 10k examples for validation. I have decided to ignore 3,657 examples in-between.

\paragraph{Vocabulary}

The training dataset was lowercased in order to avoid an issue of data sparsity, as the size of the dataset is relatively small. Since the provided training examples were already tokenized to a certain degree, I have not attempted any further tokenization, other than removing a quotation mark ``"''. In the case of blind development and test sets, I used spaCy\footnote{\url{https://spacy.io/}} for automatic tokenization. At this stage, a vocabulary was built using all the $n$-gram's with $n$ up to 2 from the entire training set. This resulted in a vocabulary of 102,608 unique $n$-gram's, and among them, I decided to use only the 100k most frequent $n$-grams. 

\section{Model and Training}

The ``strawman'' is an ensemble of five deep bag-of-ngrams classifiers. Each classifier is a multilayer perceptron consisting of an embedding layer which transforms one-hot vector representations of words into continuous vectors, averaging pooling, a 32-dim $\tanh$ hidden layer and a binary softmax layer. The classifier is trained to minimize cross-entropy loss using Adam~\citep{kingma2014adam} with the default parameters. Each training run was early-stopped based on the validation accuracy and took approximately 10-20 minutes on the author's laptop which has a 2.2 GHz Intel Core i7 (8 cores) and does not have any GPU compute capability. The output distributions of all the five classifiers, which were initialized using distinct random seeds, were averaged to form an ensemble. The entire code was written in Python using PyTorch.\footnote{\url{http://pytorch.org/}} The implementation is publicly available at \url{https://github.com/kyunghyuncho/strawman}.

\section{Result and Thoughts}

Despite its simplicity and computational efficiency, the ``strawman'' fared reasonably well. The ``strawman'' was ranked first in terms of the average F1 score on all the breakers' test cases, outperforming more sophisticated systems based on a recursive deep network (Builder Team 5, \citep{socher2013recursive}) as well as a convolutional network (Builder Team 6, \citep{kalchbrenner2014convolutional}). When measured by the proportion of the test cases on which the system was broken (i.e., the system is correct only for one of the minimally difference sentences and wrong for the other), the ``strawman'' was ranked fourth out of six submissions, although the margin between the ``strawman'' and the best ranking system (Builder Team 2) was only about 1\% out of 25.43\% broken case rate, corresponding to 6 cases. 

Although we must wait until the breakers' reports in order to understand better how those broken cases were generated, there are a few clear holes in the proposed ``strawman''. First, if any word is replaced so that a new bigram disappears from the predefined vocabulary of $n$-grams, the ``strawman'' could easily be thrown off. This could be addressed by character-level modelling~\citep{ling2015finding,kim2015character} or a hybrid model~\citep{miyamoto2016gated}. Second, the ``strawman'' will be easily fooled by any non-compositional expression that spans more than two words. This is inevitable, as any expression longer than two words could only be viewed as a composition of multiple uni- and bi-grams. Third, the obvious pitfall of the ``strawman'' is that it was trained solely on the provided training set consisting of less than 7k full sentences. The ``strawman'' would only generalize up to a certain degree to any expression not present in the training set.

\section*{Acknowledgments}

The author thanks support by eBay, TenCent, Facebook, Google and NVIDIA. 


\bibliography{emnlp2017}
\bibliographystyle{emnlp_natbib}

\end{document}